# P(Yes): 15× Improvements via Test-Time Scaling of Decision Models

**Lev Stambler**
Tear Labs
University of Maryland, College Park
lev@tearlabs.ai

**Abstract**

Decision models answer typed questions, such as multiple-choice or yes/no questions, with a probability for each option. We add test-time search to frozen decision models, with no training: we show the model a complete proposed answer, read off its probability of saying "Yes" to the proposal's correctness, and search for answers that it rates higher. The search is done either by editing one answer at a time or by joint optimization, which changes several answers at once. Search raises accuracy for every model and test set we tried, and the gains are largest for exact correctness on logic puzzles, where every answer in a task must be right. On 128 such puzzles, it takes Qwen3.8-27B from 2 to 30 exactly correct puzzles (15×) and its per-blank accuracy from 27.8% to 48.5%. On a shared set of 47 puzzles, searching over each model's own Yes/No score raises the per-blank accuracy of StartLux Instant, Perplexity Decider, Clef, and JEV, which we reach only through its public API, by 4.0 to 13.0 percentage points, and raises the number of puzzles they get exactly correct, out of 47 and averaged over three search seeds, from 10 to 35, 9 to 26, 2 to 11, and 9 to 31.7 respectively.

## 1 Introduction

Decision models are language models built to make typed decisions: given some context and a question with a fixed list of options, they return a probability for each option [10]. Models such as StartLux Decision [33], Perplexity Decider [26], Cloudflare Clef [4], and JEV are trained for exactly this, and a general-purpose model such as Qwen [36] can be used the same way. Many real tasks, however, consist of several such choices that depend on each other, such as the fields of a form or the seats in a seating plan, and answering each choice on its own can produce answers that contradict each other [14, 22]. We report two measures on these tasks: *per-blank accuracy*, the share of individual blanks that are right, and *exact correctness*, the share of tasks where every blank is right.

The two measures reward different things. Per-blank accuracy goes up whenever a single blank is fixed, so a model that answers each blank on its own already does reasonably well by it. Exact correctness, in contrast, gives no credit until the last wrong blank is fixed, and when blanks constrain each other, the last few are often the ones that can only be fixed together.

This is where a model's judgment of a complete answer helps most, since it sees all the blanks at once. We therefore expect search over complete answers to help most with exact correctness while still raising per-blank accuracy along the way, and this is what we find.

To see why this is hard for a decision model, consider a puzzle that seats four guests, Ana, Ben, Cleo, and Dev, in seats 1 to 4. Every guest needs a seat, no two guests can share one, and a list of clues such as "Dev sits somewhere left of Ana" has to hold for the arrangement as a whole. A decision model that answers each guest's question on its own can make four choices that each look reasonable and still put two guests in seat 2, which makes the whole seating wrong. Notably, this can happen even when the same model, shown a complete seating, can tell whether that seating is correct. This raises the question we study in this paper: can a decision model's judgment of complete answers guide a search for better ones? We use the puzzle in Figure 1 as a running example throughout.

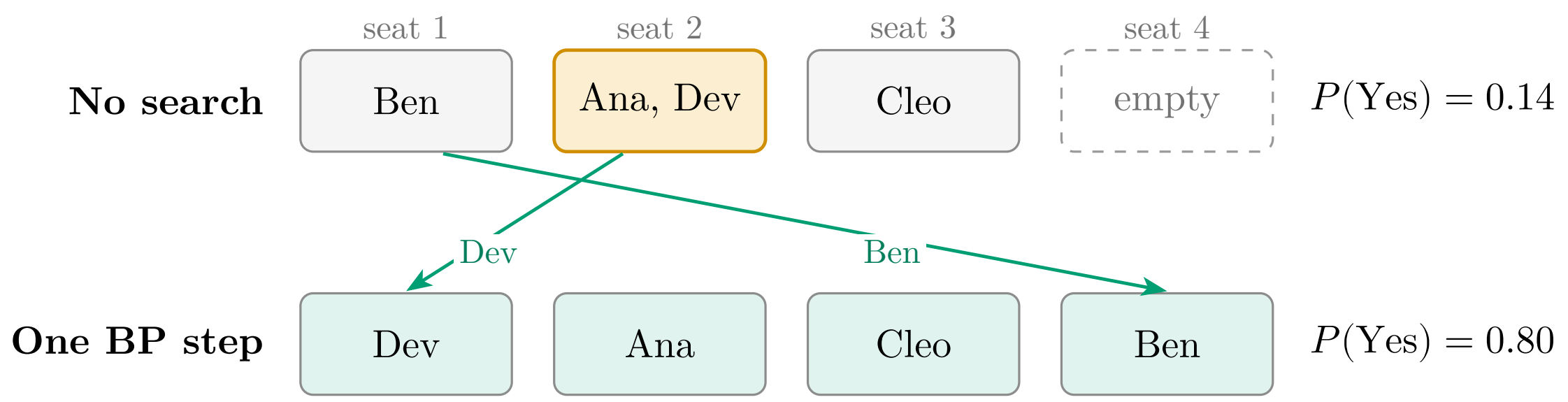

**Figure 1.** The running example, held-out seating puzzle 24297. The no-search answer puts Ana and Dev in seat 2 and Ben in seat 1, and the model scores it at $P(\text{Yes}) = 0.14$. Moving Ben to seat 4 and Dev to seat 1 gives the unique solution, which scores 0.80. Either move alone still leaves a conflict. One BP step finds the solution after checking 37 complete assignments, including initialization.

This is the idea behind P(Yes). A complete answer to a multi-blank task can itself be judged as one more yes/no decision, so we give a frozen decision model a complete answer, ask whether it is correct, and use the probability of "Yes" as a score: we change the answer, check it again with the model, and keep the best answer seen so far. The model never generates a solution of its own and its weights never change, so the recipe applies to any decision model that can score a complete answer, including one that we can only call through an API. P(Yes) brings together ideas from self-evaluation [13, 40], energy-based inference [17], and test-time computation [30], and we think of it as test-time scaling for decision models.

For this to work, the model's score has to prefer a correct answer and the search has to find that answer. Since the two can fail independently, we score every possible answer on small tasks, which lets us tell score errors apart from search errors.

Search improves both measures across every model and set we tried, and the gains are largest for exact correctness on logic puzzles. Our controlled experiments use open Qwen models, since open weights let us measure every model evaluation and runtime directly. On 128 synthetic logic puzzles, search raises Qwen3.8-27B's per-blank accuracy from 27.8% to 48.5% and its exactly correct puzzles from 2 to 30, 15× its initialization, although most puzzles are still wrong. On a shared set of 47 puzzles, the same search over each model's own Yes/No judgment raises the per-blank accuracy of StartLux Instant, Perplexity Decider, Clef, and JEV by 4.0 to 13.0 percentage points, and raises the number of puzzles they get exactly correct, out of 47 and averaged over three search seeds, from 10 to 35, 9 to 26, 2 to 11, and 9 to 31.7 respectively, without updating any weights (Section 4.4). On two published sets of short multi-field forms (Appendix A), the gains are smaller and depend on the set. Finally, Section 5 places our results next to JEV's out-of-the-box answers as a reference point, although not at equal compute, since JEV's internal compute is unknown.

# 2 Our Scoring Method

Before moving on to our search methods, we first describe how we score each completed answer. We will define both our scoring method and a way of counting how many answers a search has checked. The score asks only for the model's probability of one option in one yes/no decision, which every decision model provides, so we describe it here for a general model and give the details of each trained decision model's reader in Section 4.4.

**Search Space:** We call a filled-in answer to a task an *assignment*, where every blank gets one of its allowed answers. For example, $(2, 4, 3, 1)$ seats Ana, Ben, Cleo, and Dev in that order. If blank $i$ has candidate set $C_i$, the possible assignments are the product $C_1 \times \cdots \times C_n$, and we write $x$ for a single assignment. Only $x$ changes during a search, while the task and its allowed answers stay fixed. To score an assignment, we fill its candidates into the form and append a fixed verification question:

> *Is the entire completed fill correct for every blank, given the context and assignment instructions? Answer only " Yes" or " No".*

**Reading the Score:** We give the decision model the task, the proposed solution, and this question, and we take its probability of "Yes". A model trained for decisions returns this probability directly. For the Qwen models, which are general-purpose language models, the prompt ends at `Answer:`, and we take the next-token logits $z_{\text{Yes}}$ and $z_{\text{No}}$ for " Yes" and " No" and apply a softmax over just these two tokens. Note that the model only scores the answer we give it and never writes a solution or any reasoning of its own.

**Energy:** Writing $p(x)$ for an assignment's Yes score, we define its energy as

$$E(x) = -\log p(x), \tag{1}$$

so that maximizing the score is the same as minimizing the energy [17]. For Qwen, $p(x) = \exp(z_{\text{Yes}})/(\exp(z_{\text{Yes}}) + \exp(z_{\text{No}}))$. Because this probability is only over "Yes" and "No", for Qwen and for the trained decision models alike, we use it to rank answers and not as a calibrated chance of being correct. A score of 0.80, for instance, does not mean an 80% chance of being right.

**Counting Evaluations:** We measure the cost of a search in NFE, the *number of function evaluations*, where one NFE is the model scoring one distinct complete assignment. So changing one seat or all four seats costs the same single NFE, and ten distinct assignments cost ten NFE even if we score them in one GPU batch. If a search comes back to an assignment it has already checked, we reuse the cached score for free. NFE counts every complete assignment checked during initialization and search, but it leaves out the shared-prefix computation, the soft-input initialization below, and BP's arithmetic, which is why we also report runtime, which includes all of them. We give both as averages per task, and we measure runtime within batches, not as the latency of a single request.

**Reusing the Shared Context:** All proposals for a task share the same clues, instructions, and allowed answers, so we cache that prefix once and only compute the solution and verification question for each new proposal. Every score still sees the full task context.

**Initialization (No Search):** To get a starting point, we first rank each blank's candidates while treating every other blank as a uniform mixture of its candidate embeddings[code]. We do not count these soft-input evaluations in NFE. We then use the rankings to build a shortlist of complete assignments, score each one, and keep the best, which takes about 13–16 NFE per task on average across the three sets. Throughout the paper, "no search" means stopping here, before any of the search methods in Section 3.

**Checking Every Answer:** When a task has at most 2,048 assignments, we can afford to score all of them and see whether a lowest-energy answer is exactly correct. For each error, this tells us whether the search missed a better-scoring answer or whether the score itself prefers a wrong one.

# 3 Search

The natural starting point for search is to change one blank at a time. Unfortunately, this kind of search can stall: if two guests need to swap seats, either change on its own may make the answer worse. We therefore compare it with two methods that can coordinate changes across blanks, namely BP steps and coupled flips. All three methods use the same model, score, and initialization, and none of them looks inside the model: each one only proposes complete assignments and reads the decision model's score for them, so the same search runs unchanged on any decision model, whether we host it ourselves or call it through an API.

Each method proposes complete assignments, checks them with the model, and always keeps the lowest-energy answer it has checked so far. As a result, more search never makes that score worse, although it can still lower exact correctness when the model prefers a wrong answer. Local methods that change one coordinate at a time share this basic limitation [3]. The three methods work as follows.

1. **Random switches**code. Starting from the best assignment seen so far, we choose up to four untried single-blank changes at random, score the resulting assignments, keep the best, and repeat. The search stops when it reaches its round budget or when no untried single-blank changes remain around the current best assignment.
2. **BP steps**code. We evaluate changes to individual blanks and to pairs of blanks, use them to build a local pairwise approximation to the energy, and then run belief propagation on that approximation to propose a joint assignment. We derive the approximation in Section 3.1.
3. **Coupled flips**code. We propose a random change to one blank and, holding that change in place, evaluate every candidate of every other blank. Shifts in those blanks' preferred candidates suggest follow-up moves, so we evaluate the resulting joint proposal and keep it if it lowers the energy.

All three methods are implemented in the search entry point. At larger budgets, a local search additionally evaluates random changes to one or two blanks.

## 3.1 How a BP step works

To understand what a BP step buys, it helps to first see why single-blank moves fail. Suppose two guests occupy the wrong seats, so that moving either guest alone creates a conflict while moving both resolves it. A method that only accepts improving single-blank moves can never make this joint change. In Figure 1, for instance, Ben and Dev both have to move before the seating becomes correct.

What we want, then, is a cheap approximation to the energy that captures such pair interactions. Fix the current assignment $x^{(0)}$ and let $E_0 = E\big(x^{(0)}\big)$. We write $x^{(0)}[i \to c]$ for the assignment that changes blank $i$ to candidate $c$ and leaves every other blank as it was. With this notation, we build the approximation in four steps.

**Single-blank terms.** First, we evaluate each assignment that changes a single blank, and for candidate $c$ of blank $i$ we define

$$u_i(c) = E\big(x^{(0)}[i \to c]\big) - E_0. \tag{2}$$

This is the cost of changing blank $i$ while leaving all other blanks fixed. With four guests and four seat choices, for example, there are 12 single-blank alternatives, three per guest, and we evaluate only those that have not already been scored.

**Pair terms.** Next, for each pair of blanks $i, j$, we evaluate joint changes using each blank's $k$ best candidates, with $k = 2$, and define

$$\psi_{ij}(a, b) = E\big(x^{(0)}[i \to a, j \to b]\big) - E_0 - u_i(a) - u_j(b). \tag{3}$$

The subtraction removes the part of the cost that the two single-blank terms already explain, so a negative residual means that the joint change costs less than the two separate changes added together. Pair terms that we did not measure are set to zero.

**Joint proposal.** We then use BP to look for an assignment with a low value of the approximate energy

$$\hat{E}(x) = E_0 + \sum_i u_i(x_i) + \sum_{i<j} \psi_{ij}(x_i, x_j), \tag{4}$$

whose graph, shown in Figure 2, has one node per blank and an edge for every pair. On this graph we run 32 rounds of loopy min-sum BP[code] [16, 24]. Intuitively, the messages summarize which choices at one blank work well with a given choice at another. Since this is only arithmetic on costs we have already measured, it adds no model evaluations.

**Model check.** Finally, we evaluate BP's proposal under the original energy $E$. When the domain has at most 65,536 assignments, we also evaluate an exact minimizer of $\hat{E}$, together with nearby proposals obtained by changing BP's least confident blanks, and keep whichever proposal has the lowest model energy. These nearby proposals follow ordered-statistics decoding (OSD) [8], which is also used together with BP to decode quantum LDPC codes [25, 28].

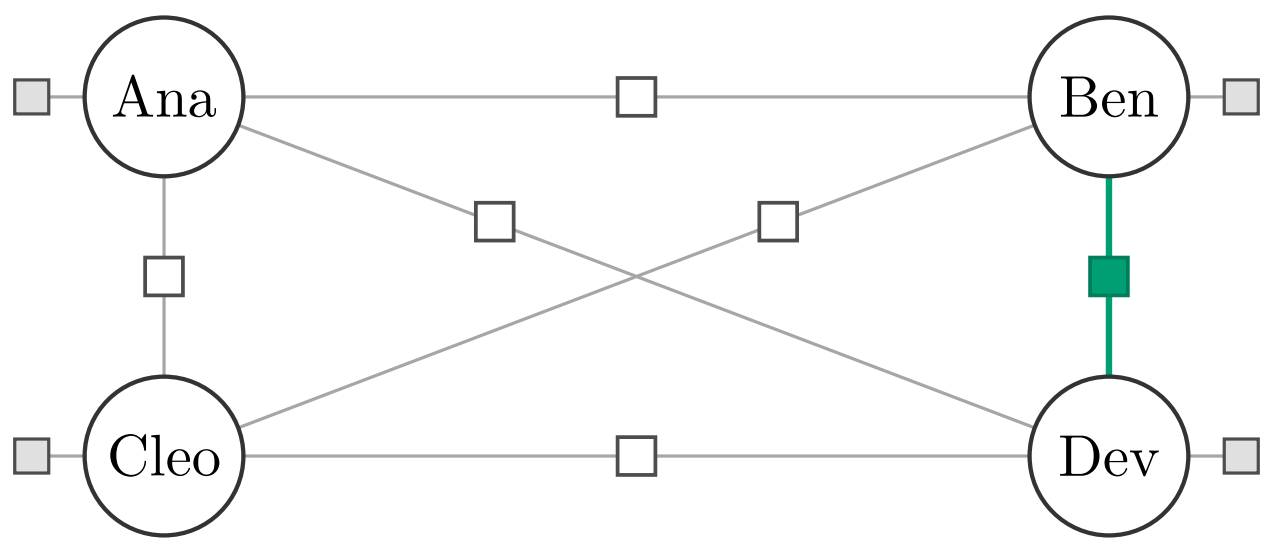

**Figure 2.** A pairwise approximation for the running example. Each node records the cost of moving one guest, and each edge records the extra effect of moving two guests together. The highlighted edge is Ben-Dev. BP uses these costs to propose a joint answer, which we then check with the model.

**Why subtract the single-blank terms?** The single-blank terms already account for the separate effects of moving $i$ and $j$, so the pair term should record only what changes when we move them together. For example, if the energy is additive, $E(x) = \text{constant} + \sum_i f_i(x_i)$, then substituting into Equation 3 gives $\psi_{ij}(a, b) = 0$. A nonzero residual therefore signals a failure of additivity at the measured assignments. It cannot, however, detect interactions among three or more blanks, which is why we still check every proposal with the original model.

**The two-blank case.** When there are only two blanks and the pair evaluations cover both full candidate sets, the definition of $\psi_{12}$ gives $\hat{E}(a,b) = E(a,b)$ for every assignment. It is not too hard to see this, since the single-blank terms cancel. In this case, whenever the exact-solve check is available, the step finds a global minimizer of the original energy, whereas loopy BP on its own has no such general guarantee.

With more blanks, the approximation leaves out interactions among three or more blanks, as well as pair interactions outside the selected top-$k$ candidates, although every candidate remains reachable through its single-blank term. Checking proposals with the model ensures that an approximation error never worsens our best score, but search still cannot find an answer that it never proposes. For four blanks, a step takes a few dozen NFE; the running example uses 37, including initialization. Repeating the step rebuilds the approximation around the new answer, and at fixed $k$ the cost of evaluating pairs grows quadratically with the number of blanks. For background on belief propagation, we refer the reader to Yedidia, Freeman and Weiss [38] or to MacKay's textbook [20].

The motivation here is similar to that of our earlier work on small-set-flip (SSF) stopping failures, where we added a complementary correction step once local progress stalled [31]. Here, though, the objective is a model score rather than a syndrome weight, so the decoding guarantees from that setting do not carry over.

# 4 Experiments

## 4.1 Test sets and models

We evaluate on three sets of tasks, which differ in where they come from and in how cleanly they test the method. All three are published with the code, and Appendix A describes each of them in detail.

- **Logic puzzles (128).** These are synthetic seating puzzles like the one in Figure 1, together with $4 \times 4$ Latin squares (filling a grid so that no symbol repeats in a row or column), for 654 blanks in total. The published code reconstructs the puzzles.
- **Dev forms (68).** These are short forms with 2–4 blanks each, where every blank is a multiple-choice or yes/no decision with a fixed list of options (`forms/dev_forms.jsonl`). Most of them (41) are spreadsheet fills derived from SpreadsheetBench, and the rest are code, computer-use, safety, and trace-quality decisions that we wrote for this project. Every task in this set had already been selected by earlier experiments, so it is not a clean test.[1]
- **Held-out forms (68).** These are forms of the same kind that no earlier experiment had used, all written for this project: 40 agent-trace quality judgments, 23 code-compression decisions, and 5 computer-use actions (`forms/heldout_forms.jsonl`). This makes them the fairer test on forms.

Both form sets come from the same frozen pool of typed-decision tasks and are selected by the same rule, with opposite answers to whether an earlier experiment had already used a task (Appendix A.2). For the controlled experiments we use frozen Qwen3.8-27B and Qwen3.5-9B [36] as decision models, both run in BF16 on a single B200. Since their weights are open, we can count every model evaluation, measure runtime, and score every answer on small tasks, which a hosted model does not allow. Section 4.4 then runs the same search on four models that were trained specifically for decisions.

[1]We call this set "dev" because it informed development. Each of its tasks had been selected by earlier experiments in this project, and those experiments reported results on it, so choices we made during development may reflect it. Its numbers are therefore development evidence rather than a held-out test. The held-out forms, in contrast, were untouched until 2026-09-23, although they have since been used repeatedly as a confirmation set; see Appendix A.2.

## 4.2 Score quality

We begin with the exhaustive check from Section 2, whose results appear in Table 1. Each row uses the same tasks in every column, namely those with at most 2,048 assignments, matched across both models. On dev forms, checking every answer beats initialization, which means there is room for search to help. On the 51 small logic puzzles, however, the 27B's best-scoring answer is exactly correct only 41.2% of the time, so even finding the exact optimum of the score would leave many errors. Since these subsets need not represent the full sets, their results should not be read as ceilings on full-set exact correctness.

**Table 1.** Exhaustive selection and initialization on identical enumerated cohorts (at most 2,048 assignments). Each cell gives exact correctness and correct/total tasks.

| Set / matched tasks | 27B exhaustive | 27B init. | 9B exhaustive | 9B init. |
|---|---|---|---|---|
| dev forms<br>64 tasks | 89.1%<br>57/64 | 73.4%<br>47/64 | 34.4%<br>22/64 | 17.2%<br>11/64 |
| logic puzzles<br>51 tasks | 41.2%<br>21/51 | 3.9%<br>2/51 | 23.5%<br>12/51 | 9.8%<br>5/51 |
| held-out forms<br>68 tasks | 48.5%<br>33/68 | 45.6%<br>31/68 | 8.8%<br>6/68 | 8.8%<br>6/68 |

## 4.3 What extra NFE buy

With the score and the three searches in place, we can now ask what the extra evaluations buy. Figure 3 and Figure 4 plot each search's exact correctness as it evaluates more proposed solutions.

The benefit of extra compute depends on the task. On logic puzzles, where coordinated changes matter more, BP and coupled flips keep improving after random switches have stopped. At our largest budget, one BP step followed by 32 rounds of random one- and two-blank changes reaches 30/128 (23.4%) at about 217 NFE and 1.8 seconds per task. On dev forms, random switches and one BP step reach 86.8% at roughly 21 NFE, and coupled flips improve further with their joint proposals, reaching 89.7%. On held-out forms, search gives a smaller gain, from 45.6% to 50.0%. Since exhaustive selection is exactly correct only 48.5% of the time on the enumerated subset of these forms, this suggests that the score itself needs work, although it does not establish a ceiling for the full set.

Random switching stops below 50 NFE because it runs out of moves. We ran it for up to 64 rounds, and on the puzzles both the 32- and 64-round runs average 43 NFE and are exactly correct on 8/128 tasks: once the search has exhausted the unchecked single-blank moves around its best answer, raising the round limit does almost nothing. The "BP, then random" curve can keep going because it also tries two-blank moves. We have not tested random restarts or simulated annealing [15].

Search raises per-blank accuracy as well, the share of individual blanks that are right, although by less than it raises exact correctness. Table 2 uses the same runs as Table 5, comparing no search with each Qwen model's best search setting. For the 27B, per-blank accuracy goes from 85.9% to 92.9% on dev forms, from 64.3% to 68.3% on held-out forms, and from 27.8% to 48.5% on the logic puzzles, and the 9B improves on all three sets as well, so the gains in whole answers come with gains in individual blanks. JEV out of the box still gets more puzzle blanks right (70.5%), even though fewer of its puzzles are exactly correct. We have no per-blank count for JEV on the held-out forms.

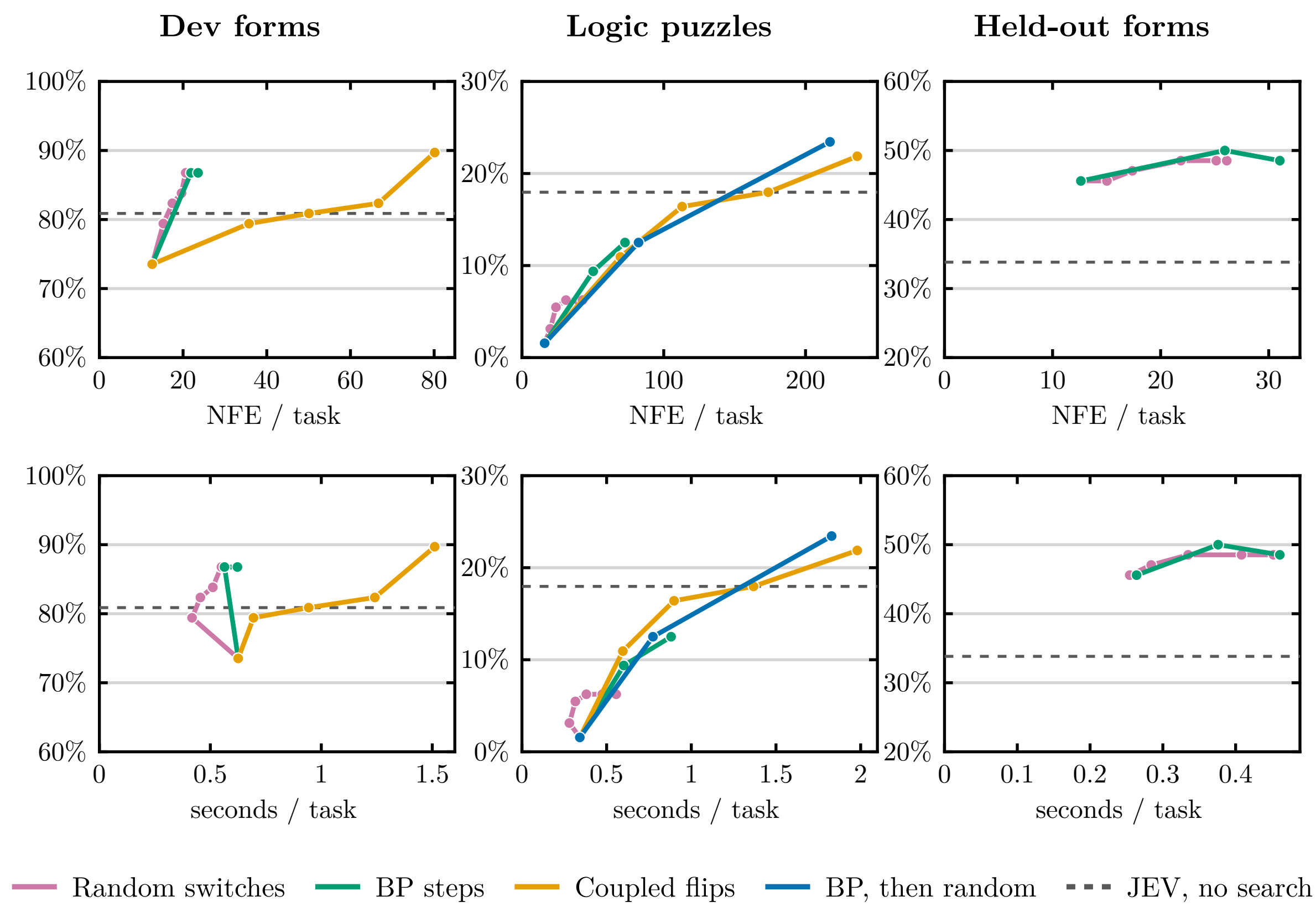

**Figure 3.** Qwen3.8-27B: exact correctness (every blank right), against NFE per task (top) and seconds per task on one B200 (bottom). NFE includes the initialization shortlist, and runtime includes the full procedure. Every curve starts at the no-search baseline. For reference, the dashed line is JEV's exact correctness through its API without search; its internal NFE and comparable runtime are unknown, so it does not mark an equal-compute comparison. We did not run coupled flips on held-out forms.

**Table 2.** Per-blank accuracy (correct blanks over all blanks) for the runs in Table 5. JEV runs through its API without search; we have no per-blank count for it on the held-out forms.

| Set | JEV API | 27B init. | 27B search | 9B init. | 9B search |
|---|---|---|---|---|---|
| dev forms | 85.9% | 85.9% | **92.9%** | 47.0% | 66.7% |
| held-out forms | — | 64.3% | **68.3%** | 42.1% | 43.9% |
| logic puzzles | **70.5%** | 27.8% | 48.5% | 37.2% | 42.8% |

## 4.4 Results across decision models

We first compare the two Qwen models, which shows that a good starting point and a good score need not go together. Without search, the 27B is exactly correct on more dev forms than the 9B (73.5% versus 16.2%) and on more held-out forms (45.6% versus 8.8%). Its highest-scoring answer is also exactly correct more often, both on the enumerated dev forms (89.1% versus 34.4%) and on the enumerated held-out forms (48.5% versus 8.8%).

The puzzles, however, go differently. Here the 9B starts slightly ahead without search (4.7% versus 1.6%), but the 27B finishes ahead with search (23.4% versus 13.3%), and on the enumerated puzzles the 27B also selects the exactly correct answer more often (41.2% versus 23.5%). The quality of the starting solution and the quality of the verification score therefore need not agree. Since the two models differ in release as well as in size, we cannot attribute this difference to parameter count alone.

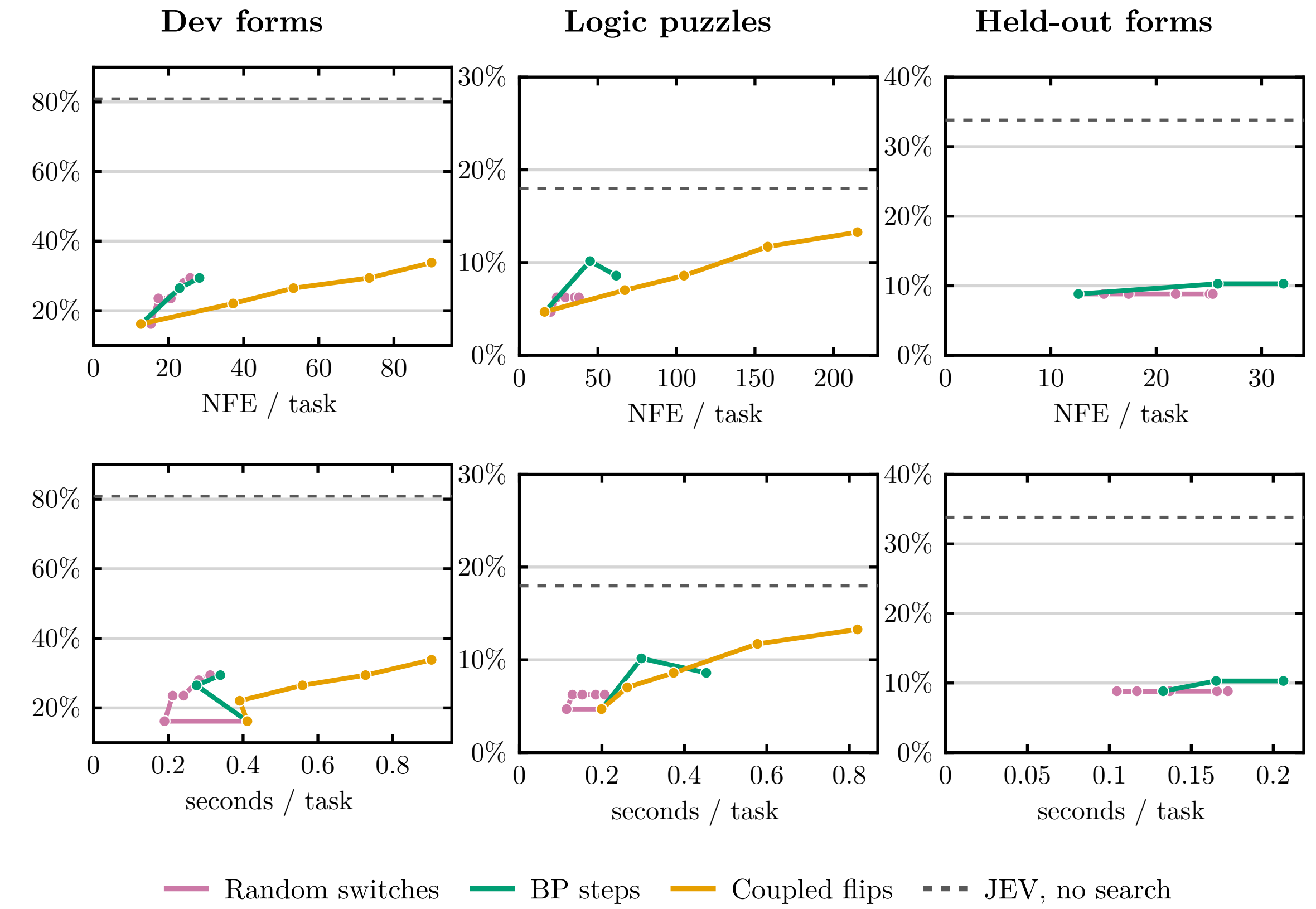

**Figure 4.** Qwen3.5-9B, laid out as in Figure 3. We did not run coupled flips on held-out forms.

Beyond the two Qwen models, search also raises the exact correctness of frozen 27B models that were trained specifically for decisions, namely StartLux Instant v1.1, Perplexity Decider v1.1, and Cloudflare Clef. It also raises the exact correctness of JEV, which we can reach only through its public API, since JEV can still judge a complete fill and so its judgment can drive the same search. For this comparison we use the first 47 logic puzzles completed by all four open search models, which include a Qwen3.8-27B control, and run JEV on the same puzzles. Each model's native decisions initialize the search, and its own binary judgment of a complete fill supplies the energy. We then apply measured-factor BP, an OSD shortlist, and local refinement, with a cap of 256 additional complete-assignment evaluations, and average the search results over three seeds.

For every proposed fill, the same model answers the question "Is every assigned answer correct, given the context and questions?", and search minimizes $-\log P(\text{yes})$. Clef, Perplexity, and StartLux use their native decision probabilities, with Perplexity keeping its released temperature calibration, while Qwen uses probabilities normalized over the no/yes answer-letter tokens. These scores only rank candidates within each model, whereas exact correctness compares every answer with the gold answers, which search never sees. Throughout, each checkpoint keeps its native inference format, decision head, attention behavior, and calibration, and no model weights are updated.

JEV (`typesafe/jev-1.13`) is a hosted model, so we run the same arms, seeds, and cap through its public decisions API. Its native answers initialize the search, and for each fill we send it the same question as a single binary decision and use its returned probability of "yes". Two properties of the API matter here. First, JEV returns probabilities rounded to two decimals, so many fills tie and BP sees costs only in steps of 0.01, and when the best energy ties, search keeps the first fill it found. Second, the API is not fully deterministic, so JEV's native answers on a few of these puzzles differ from an earlier API run on the same tasks. Since JEV's times are API wall-clock times rather than B200 seconds, we do not compare them with those of the open models.

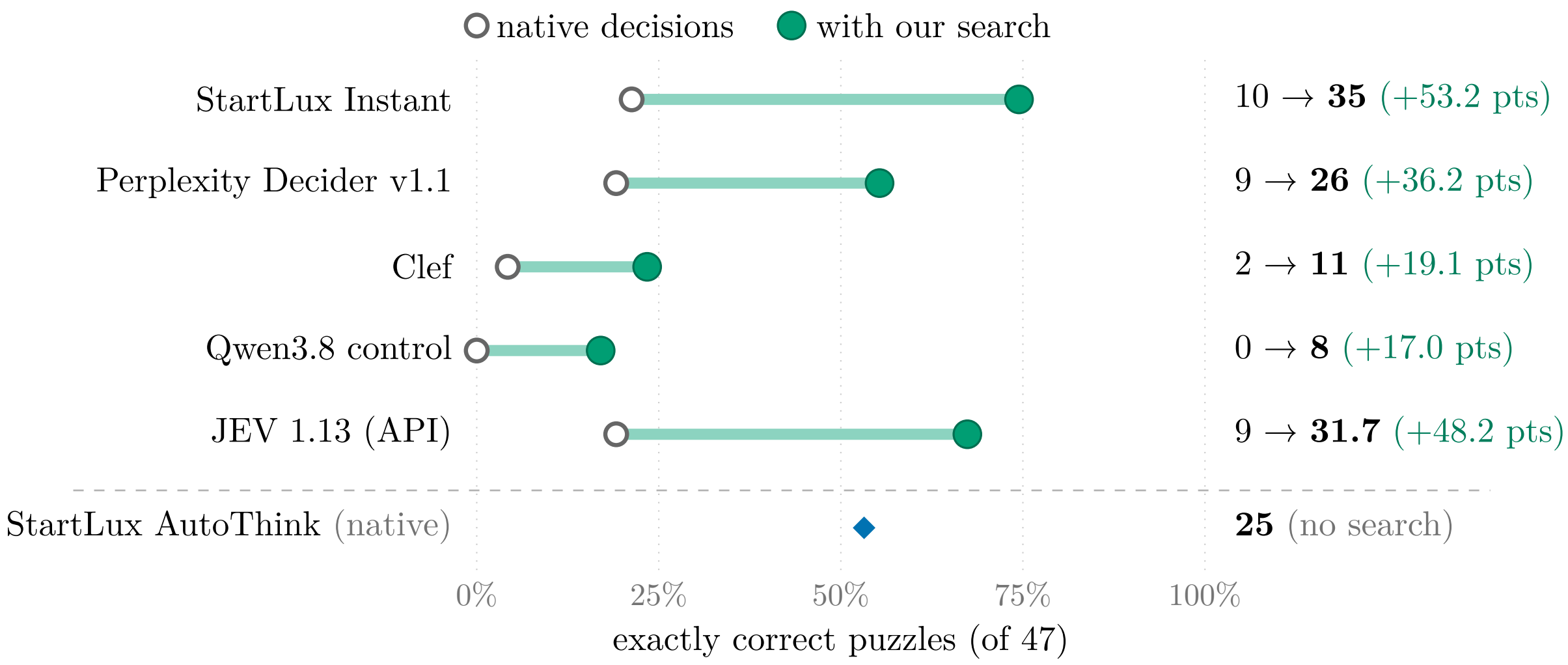

**Figure 5.** Native decisions versus our search on the same 47 logic puzzles. The open models are 27B, and JEV runs through its public API. Open circles show native decisions, and filled circles show BP and refinement search, averaged over three seeds with a cap of 256 additional complete-assignment evaluations. Gains are percentage points, and labels give exactly correct puzzles. The diamond shows native StartLux AutoThink in one run, without our search. This is a paired comparison of exact correctness, and actual compute differs.

As Figure 5 and Table 3 show, StartLux Instant improves from 10 to 35/47, a gain of 53.2 percentage points in exact correctness, and Perplexity and Clef improve as well. Search counts here are averages over three seeds, which is why some of them are fractional. JEV goes from 9 to 31.7/47, a gain of 48.2 percentage points in exact correctness (95% bootstrap interval 33.3 to 62.4), compared with 16.7 puzzles for random edits. In other words, decision training still leaves room for search over the model's own judgments, even for a model that we can only call through an API. Note that the Qwen control here uses discrete option selection rather than the soft-input initializer of the main experiment, so its initialization counts are not directly comparable across experiments.

**Table 3.** Exactly correct puzzles out of 47. Search columns average three seeds, with a cap of 256 additional complete-assignment evaluations.

| Model | Native | Random edits | BP + refinement |
|---|---|---|---|
| StartLux Instant | 10 | 18.3 | **35** |
| Perplexity Decider v1.1 | 9 | 16.3 | **26** |
| Clef | 2 | 2.7 | **11** |
| Qwen3.8 control | 0 | 0.3 | **8** |
| JEV 1.13 (API) | 9 | 16.7 | **31.7** |

As Table 4 shows, BP and refinement also raise the per-blank accuracy of all five models, by 4.0 to 13.0 percentage points. JEV, for instance, goes from 71.3% to 83.9% of blanks right, and only Clef's 95% interval reaches below zero, with a lower end of -3.5 percentage points. Random edits move the other way: they lower the per-blank accuracy of every model even as they raise its exact correctness, so on these puzzles only the coordinated proposals improve both measures.

**Table 4.** Per-blank accuracy on the same 47 puzzles and runs as Table 3, pooled over all blanks. Search columns average three seeds; the last column gives the BP and refinement gain in percentage points with a paired bootstrap 95% interval over puzzles.

| Model | Native | Random edits | BP + refinement | BP gain [95% CI] |
|---|---|---|---|---|
| StartLux Instant | 75.7% | 65.0% | **88.7%** | +13.0 [7.5, 18.4] |
| Perplexity Decider v1.1 | 73.7% | 64.4% | **81.2%** | +7.6 [1.5, 14.3] |
| Clef | 47.4% | 32.9% | **51.4%** | +4.0 [-3.5, 11.9] |
| Qwen3.8 control | 38.1% | 28.5% | **49.4%** | +11.3 [3.9, 18.0] |
| JEV 1.13 (API) | 71.3% | 64.0% | **83.9%** | +12.7 [7.3, 18.0] |

StartLux AutoThink, which already spends test-time compute on generated reasoning, offers a useful point of comparison. It is exactly correct on 25/47 puzzles at 25.6 seconds per puzzle, compared with 35/47 at 12.9 seconds for Instant with our search. Both are BF16 measurements on one B200 in the reference runner rather than optimized serving latencies. The comparison is between two ways of scaling at test time, and we have not added our search on top of AutoThink.

Finally, the random control often stops early: for StartLux, random edits use about 26 additional evaluations, compared with 201 for BP and refinement, so Table 3 does not establish an advantage over random search at matched compute. We have also not yet run broader decision tasks or cross-model Sudoku, so these results support transfer on these logic puzzles rather than a gain across all decision tasks.

# 5 JEV as a decision model

JEV (`typesafe/jev-1.13`) is a hosted decision model that we can only reach through its public API, which makes it a useful test of how general the approach is. Since the API can judge a complete fill, we can search over JEV's own Yes/No score exactly as we do for the open models, with no access to its weights or internals. On the 47 paired logic puzzles of Section 4.4, this search takes JEV from 9 to 31.7 exactly correct puzzles, a gain of 48.2 percentage points in exact correctness (95% bootstrap interval 33.3 to 62.4), compared with 16.7 puzzles for random edits. So search helps JEV in the same way that it helps the other decision models, although we have not yet run it over JEV on the forms.

For reference, Table 5 also lists JEV's out-of-the-box exact correctness on all three of our sets, next to Qwen with and without search. These numbers show where search puts an open model relative to a strong hosted decision model used as it comes, but they are not at equal compute, since we do not know JEV's internal NFE or a comparable runtime.

**Table 5.** Exactly correct tasks for JEV used as it comes, without search, and for Qwen at its best measured search setting. JEV's internal compute is unknown, so these are reference numbers rather than matched-compute comparisons.

| Set | JEV API | 27B init. | 27B search | 9B search |
|---|---|---|---|---|
| dev forms | 55/68 | 50/68 | **61/68** | 23/68 |
| held-out forms | 23/68 | 31/68 | **34/68** | 8/68 |
| logic puzzles | 23/128 | 2/128 | **30/128** | 17/128 |

With search, the 27B is exactly correct on more tasks than out-of-the-box JEV on all three sets. On the puzzles, for instance, Qwen's initialization checks 16 complete assignments, and our largest search checks about 217 NFE per puzzle (201 additional checks), taking 1.8 seconds per puzzle on one B200. This search is exactly correct on 30/128 puzzles (23.4%), compared with 23/128 (18.0%) for JEV without search. The 9B, in contrast, stays below JEV on all evaluated sets.

The comparison also shows why we focus on exact correctness. JEV has higher per-blank accuracy on the puzzles (70.5% versus 48.5%), even though fewer of its puzzles are exactly correct, and a puzzle or form with one wrong blank is still wrong. On JevBench's held-out families, by contrast, every task has a single blank, so there are no joint choices for search to refine and the two measures coincide. There, the 27B gets 79.3% on 82 tasks, compared with 76.8% for JEV.

# 6 Related work

**Models judging answers.** Kadavath et al. [13] study $P(\text{True})$, in which a language model is asked to estimate whether a proposed answer is correct. Our Yes/No score applies the same idea to complete structured answers that a search algorithm supplies. We use the score to rank assignments without any additional training, and we do not assume that it is a calibrated probability of correctness. Verifiers can also be trained to read off a "Yes" token [40], and models are now widely used as judges [42].

**Joint decisions and consistency.** Most similar to our work is ConCoRD [22], which combines answer likelihoods with pairwise compatibility scores from a separate natural-language-inference model and then selects answers jointly using a factor graph and weighted MaxSAT. BeliefBank [14] similarly revises stored beliefs with weighted MaxSAT and feeds them back into later queries. Both improve predictions without retraining the underlying language model. In contrast, we search the supplied finite candidate domains of a single task, and we measure interactions by changing its completed assignment and reading the same model's Yes/No score. BP proposes answers, while the original complete-assignment score decides which of them to keep.

**Test-time compute for typed decisions.** SanSi [10] is especially close to us in motivation. It produces typed option probabilities without generating tokens and improves its decisions by repeatedly applying the same layers, training adapters and a readout with supervision at each loop. Our search instead leaves the verifier fixed and spends compute on alternative assignments outside the model. These are two different ways of spending extra inference compute, hidden-state refinement on the one hand and explicit decision search on the other, and we have not evaluated their combination.

**Test-time compute.** Much work spends extra inference compute on sampling and checking free-form answers, whether through generated reasoning [35], verifiers that rank samples [5, 19], majority votes [34], tree search over reasoning steps [37], or choosing among these strategies for a fixed budget [23, 30]. Sample, Scrutinize and Scale [41] studies sampling with self-verification, and in particular how extra verification compute improves answer selection. Our extra compute instead buys evaluations of edits to finite assignments, without generating any reasoning traces. Since the costs of model evaluations and runtime differ across methods, we report both.

**Energy-based structured prediction.** Treating $-\log P(\text{Yes})$ as an energy follows the energy-based view of learning [17], and single-coordinate descent on such an energy goes back to iterated conditional modes [3]. Structured Prediction Energy Networks [2] already make decisions by iteratively optimizing a learned energy over candidate labels, including the dependencies between them, and Energy-Based Transformers likewise learn an energy for optimization at inference time [11]. We instead use a frozen language model's existing verification score, and we compare discrete search with relaxed gradient optimization (Appendix C.1).

**Search with frozen language models.** Training-free energy-based inference also predates this work. Mix and Match [21] combines the scores of pretrained models and uses Metropolis–Hastings sampling for controlled text generation, while COLD Decoding [27] runs Langevin dynamics over a continuous relaxation of text to satisfy generation constraints. These methods search over text sequences, whereas we search over supplied blank options and keep the best checked assignment under a single complete-answer verification score.

**Limits of verification.** Yu et al. [39] show how imperfect verifiers can misrank candidates and prune valid reasoning paths as search grows. Our exhaustive small-domain checks expose a related distinction. Failing to find the best-scoring assignment is a search error, while the best-scoring assignment being wrong is a score error, and more search cannot by itself repair the latter. For this reason, we do not claim that exact correctness rises monotonically with compute.

**Belief propagation and decoding.** BP runs on factor graphs [16] and often works well even on graphs with loops, although it is not guaranteed to converge there [24, 38]. Iterative decoding goes back to sparse parity-check codes [9]. Small-set-flip decoders correct sets of bits rather than single bits, with guarantees for quantum expander codes under expansion assumptions [7, 18], and Grospellier et al. combine them with BP [12]. Our earlier SSF+PAL work studies the structure that is left once SSF stalls and uses it to design a second correction stage [31]. Our check step follows BP+OSD decoding of quantum LDPC codes [8, 25, 28]. Unlike these decoders, however, our approximation is built from measured model energies, with no parity-check structure and no decoding guarantee.

**Gradient-guided discrete search.** Picking discrete edits with gradients goes back to HotFlip [6] and AutoPrompt [29], and it is the basis of GCG [43]. Our gradient-chosen moves (Appendix C.1) apply the same idea to blank options.

Our contribution is therefore an empirical study of test-time search for decision models, in which a frozen model's judgment of complete answers is the only signal. It compares local and joint proposals, measures how exact correctness grows with compute, and uses enumeration to separate search errors from score errors. We do not claim that optimizing model-scored answers at inference time is itself new.

# 7 Limitations

Our results have a few key limitations, the main ones being small sets with mostly single runs and a score that can prefer a wrong answer. Each set contains 47–128 tasks, and each reported setting comes from a single run, apart from the transfer study's search results, which average three seeds, so for most settings we do not estimate variation across runs. The form sets are also small and fairly narrow in what they cover, and apart from the spreadsheet tasks their labels were written for this project rather than independently validated (Appendix A).

The exhaustive evaluation covers only domains with at most 2,048 assignments. Its results diagnose the score on that subset, but they do not carry over to larger tasks without further evidence. Likewise, the pairwise BP approximation can miss higher-order interactions, and at fixed $k$ its NFE cost grows quadratically with the number of blanks, so substantially larger forms would need a sparser measurement strategy.

Most importantly, a better score can select a worse answer, and the Yes/No score is not calibrated as a probability of correctness. Our earlier work on designed transformer inputs gives a separate, worst-case reason to doubt model robustness [32], although it does not explain these particular verification errors. Finally, we have run search over JEV only on the 47 transfer puzzles, and since JEV's internal NFE is unknown, its out-of-the-box numbers in Section 5 are a reference rather than an equal-compute comparison. Its held-out-form result comes from an earlier API evaluation on the same tasks, while we measured the other two sets for this paper.

# 8 Conclusion

We have shown that a frozen decision model's own judgment can guide search over decisions that depend on each other. Spending more evaluations raises exact correctness substantially on the logic puzzles we tested, by 15× for Qwen3.8-27B used as a decision model, and the same search helps every model trained specifically for decisions that we tried, StartLux, Perplexity Decider, Clef, and JEV through its API, while the gains on forms depend on the set. These gains are not limited to whole answers, since search raises per-blank accuracy on every set and for every model as well, although the relative gains are largest for exact correctness, the stricter measure. We think of this as *test-time scaling for decision models*: given more inference compute, the model spends it checking and revising complete decisions while its weights stay fixed. In this sense, search is a way for a decision model to think, since it proposes an answer, checks it, and uses that judgment to decide what to try next.

Both the score and the search still limit the results, and a higher-scoring answer can still be wrong. This leaves a few immediate follow-up questions. Can we make the model's judgment more reliable without giving up the frozen-model setting? And how should we spend the extra compute on larger tasks, where measuring every pair becomes expensive?

Exploring different BP structures that match the problem at hand is left to future work. For decision tasks whose constraints are hard to write down, it remains open which groups of choices to connect, which checks to ask the model for, and how to balance their cost against exact correctness. When exact constraints are available, on the other hand, a rules-only control is essential, since a learned score may not be needed at all. While we do not have immediate answers to these questions, we believe that they are interesting directions for future work, and we will add Sudoku completion results in a future revision.

# Acknowledgments and AI use

AI tools helped write the code for this project and helped adapt the accompanying blog post into this paper. The ideas are the author's, with some suggestions from AI. The author reviewed the text and takes full responsibility for its content.

# Reproducibility

The published code contains the prompt, cached reader, search procedures, BP solver, and puzzle generator, in about 2,800 lines of Python. Its evaluation loop records chosen assignments, correctness, scores, NFE, and runtime, and these saved results generate the plots and the main comparison tables. Code links point at commit `0e858f2`. Models are `Qwen/Qwen3.8-27B@1d4bf0f` and `Qwen/Qwen3.5-9B@c202236`, frozen, in BF16 on one B200. Times are measured per task within batches, rather than as single-request latency. For the no-search starting points in Figure 3 and Figure 4, we use the fastest baseline timing across the corresponding search runs to reduce the effect of initial kernel compilation, and the remaining points use their recorded run times. The unpublished experiment and the prompt-rewriting results in the appendix are not part of the published code.

# A Test sets in detail

This appendix describes each test set in enough detail to rebuild it. All three sets are part of the published code at commit `0e858f2`, so every task, gold answer, and option list in the paper can be inspected directly.

## A.1 Logic puzzles

The logic puzzles are generated by the published puzzle builder. It produces 30,000 puzzles with seed 53, half seating puzzles like Figure 1 and half $4 \times 4$ Latin squares, and checks by brute force that every puzzle has a unique joint solution while every blank is ambiguous on its own. A fixed 6% of the puzzles, chosen by a hash of the job index, is held out, and the evaluation draws a fixed subset of 128 of them with seed 17. These 128 puzzles contain 654 blanks in total.

## A.2 Where the form sets come from

Both form sets are drawn from claim-eval-v2, a pool of 5,657 typed-decision tasks with 9,791 answer fields that we froze before evaluating any model on it. Each task in the pool is a short form whose blanks are typed decisions: a multiple-choice question with named options, or a yes/no question. Every task also carries a split (development, validation, or holdout) and a prior-exposure flag, which is set when an earlier experiment in this project had already selected the task or another task in its scenario cluster.

The two sets use the same selection rule with opposite exposure flags. We take the development and validation tasks with two to four blanks, in pool order. The **held-out forms** are every such task without prior exposure, which gives exactly 68 tasks, and the **dev forms** are the first 68 such tasks with prior exposure. Table 6 summarizes both sets, and the published provenance files (`forms/dev_forms.provenance.jsonl` and `forms/heldout_forms.provenance.jsonl`) record, for every task, its category, source, split, exposure flag, and label provenance.

**Table 6.** Composition of the two published form sets, computed from the files in the published code.

| | **Dev forms** | **Held-out forms** |
|---|---|---|
| Tasks | 68 | 68 |
| Blanks | 198 | 221 |
| Tasks with 2 / 3 / 4 blanks | 26 / 22 / 20 | 23 / 5 / 40 |
| Options per blank | 2–10 | 2–10 |
| Complete assignments per task | 6–2401 | 6–700 |
| Tasks with at most 2,048 assignments | 64 | 68 |
| Tasks with prior exposure | 68 | 0 |
| *Tasks by category* | | |
| Spreadsheet fill (SpreadsheetBench) | 41 | 0 |
| Code compression | 10 | 23 |
| Computer-use action | 10 | 5 |
| Safety decision | 6 | 0 |
| Agent-trace quality | 1 | 40 |
| *Tasks by label source* | | |
| Derived from SpreadsheetBench's golden workbooks | 41 | 0 |
| Written for this project | 27 | 68 |

**Held-out forms.** All 68 held-out tasks were written for this project. The agent-trace quality tasks ask for judgments about an agent's recorded trajectory. The code-compression tasks show an original snippet and candidate rewrites under a stated contract, and ask which candidate is the shortest semantics-preserving replacement and whether a given candidate is equivalent to the original. The computer-use tasks ask which action to take in a described interface state. Since every held-out task has at most 700 complete assignments, the whole set is enumerable, which is why Table 1 covers all of it. The set was untouched until 2026-09-23, when we first used it to confirm a result. We have used it repeatedly as a confirmation set since then, so it is the fairer test on forms, although it is not a pristine holdout.

**Dev forms.** Most dev tasks (41) are spreadsheet fills derived from SpreadsheetBench, reformulated so that each blank is a cell value chosen from a finite list of candidates rather than free workbook editing. The remaining tasks were written for this project and cover code compression, computer-use actions, safety decisions, and trace quality. Every dev task had been selected by earlier experiments in this project, and those experiments reported results on this set, so choices we made during development may reflect it. We therefore treat its numbers as development evidence. Four of its tasks have more than 2,048 complete assignments, so Table 1 covers 64 of them.

**Labels and scoring.** Gold answers are the pool's expected values. For the spreadsheet tasks they come from SpreadsheetBench's expert-reviewed golden workbooks, and all other labels were written for this project without independent validation. The published evaluation loop marks a blank correct when it equals the gold answer and a task exactly correct when every one of its blanks is.

**License.** The spreadsheet tasks are derived from SpreadsheetBench and are distributed under its CC BY-SA 4.0 license. Everything else in the form sets is covered by the repository's Apache 2.0 license.

### A.3 JevBench held-out families

The single-blank comparison in Section 5 uses the public portion of JevBench v1.2.2 (MIT license), which has 48 easy, 72 standard, and 111 hard cases. We sort its task families across all three tiers and hold out every third family, which leaves 82 held-out tasks, each with a single blank. The easy and standard cases were written by the benchmark's authors, while the hard tier was written by two language models that cross-reviewed each other's labels, so those labels are reviewed but not independent human ground truth.

### A.4 How JEV was evaluated

For the out-of-the-box reference numbers in Section 5, we called `typesafe/jev-1.13` through its public API and used each returned answer as it was, with no added search and no repeated successful calls; failed requests were retried only to recover from errors. The search over JEV's own Yes/No scores on the transfer puzzles in Section 4.4 is a separate experiment, described there. We measured the logic puzzles and the dev forms for this paper. JEV's result on the held-out forms comes from an earlier API evaluation of the same 68 tasks, which we recorded when the set was first used on 2026-09-23.

## B Code map

Every procedure in this paper links to the lines that implement it in the published code, pinned at commit `0e858f2`. Table 7 collects those links in one place, in the order in which the paper introduces each piece, so that readers can go straight from a definition in the text to its implementation.

**Table 7.** Where each part of the method lives in the published code. Every link points at the exact lines, at the pinned commit.

| Component | Source lines | Described in |
|---|---|---|
| Verification question appended to every proposal | `layout.py#L39` | Section 2 |
| Two-token Yes/No energy, Equation 1 | `cache.py#L227` | Section 2 |
| Caching the shared task prefix | `verifier.py#L116` | Section 2 |
| Scoring a proposal's solution and question | `verifier.py#L209` | Section 2 |
| Soft-input initialization (no search) | `search.py#L65` | Section 2 |
| Search entry point | `search.py#L109` | Section 3 |
| Random switches | `search.py#L184` | Section 3 |
| BP step: single-blank and pair measurements | `search.py#L229` | Section 3.1 |
| Loopy min-sum BP solver | `solver.py#L341` | Section 3.1 |
| Coupled flips | `local.py#L71` | Section 3 |
| Local search over one- and two-blank changes | `local.py#L129` | Section 3 |
| Logic-puzzle generator | `puzzles.py#L27` | Section 4 |
| Evaluation loop and saved records | `evaluate.py#L48` | Reproducibility |

# C Other things we tried

## C.1 Following the gradient

The discrete searches in the main text only ever evaluate actual candidate answers, so it is natural to ask whether the model's gradient could guide the choice of candidates instead. To make the energy differentiable, we represent each blank by a weighted mixture of its candidate embeddings, and we then try two ways of using the resulting gradient.

- **Relaxed ascent.** Each blank becomes a soft mix of its options, and we push the mix uphill on P(Yes), trying step sizes from 0.05 to 16, with and without annealing, for up to 32 steps. We then round to complete answers and score a short list of them.
- **Gradient-chosen moves.** The gradient at a real answer suggests which change to try next, as in gradient-guided token edits [6, 29, 43], and we still check every change with a cached forward pass on the complete assignment.

**Table 8.** Discrete search versus gradient ascent. NFE counts complete assignments and excludes soft-input and backward computations.

| Model / set | Init. | Ascent, init. NFE | Annealed, ≤256 fills | Search, ≤ascent NFE |
|---|---|---|---|---|
| 27B · dev forms | 50/68<br>13 NFE | 52/68<br>13 NFE | 47/68<br>92 NFE | **61/68**<br>80 NFE |
| 27B · logic puzzles | 2/128<br>16 NFE | 10/128<br>16 NFE | 23/128<br>256 NFE | **30/128**<br>217 NFE |
| 9B · dev forms | 11/68<br>13 NFE | 18/68<br>13 NFE | 18/68<br>92 NFE | **23/68**<br>90 NFE |
| 9B · logic puzzles | 6/128<br>16 NFE | 6/128<br>16 NFE | 15/128<br>256 NFE | **17/128**<br>215 NFE |

As Table 8 shows, the best discrete search beats annealed ascent on all four model and set pairs while using no more NFE, although this compares answers checked rather than total computation. Relaxed ascent can still improve the starting candidates; for example, the 9B goes from 11 to 18 exactly correct dev forms out of 68. Its soft-input optimization does not add NFE, but it does need model evaluations on soft inputs as well as backward passes. Annealed ascent with a shortlist of up to 256 assignments is exactly correct on 23 of 128 puzzles on the 27B. On dev forms, however, most 27B ascent settings reduce exact correctness. Gradient-chosen moves, finally, beat random moves on some model and set pairs and do worse on others, while taking 2.1 to 4.0× as long (Table 9).

**Table 9.** Gradient-chosen versus other moves, at eight rounds of four one- or two-blank changes. Cells show exactly correct tasks and seconds per task.

| Move selection | 27B puzzles | 27B dev | 9B puzzles | 9B dev |
|---|---|---|---|---|
| random | 7/128<br>0.51 s | 54/68<br>0.66 s | 11/128<br>0.19 s | 17/68<br>0.25 s |
| no-search favourites | 12/128<br>0.51 s | 58/68<br>0.66 s | 8/128<br>0.19 s | 18/68<br>0.25 s |
| gradient | 10/128<br>1.27 s | 53/68<br>1.39 s | 7/128<br>0.76 s | 19/68<br>0.78 s |
| half gradient | 7/128<br>0.82 s | 53/68<br>1.02 s | 8/128<br>0.35 s | 18/68<br>0.43 s |

To look at the gradient more directly, we start from the no-search assignments and check whether its suggested direction increases the weight on the correct candidate (Table 10). On dev forms it does so for only 50.5% of blanks. This measures something different from ranking complete assignments, since a model can prefer a correct answer and still give unhelpful gradients between mixtures of answers.

**Table 10.** Gradient direction at initialization: how often it increases the weight on the correct candidate, and how often it fixes a wrong starting candidate.

| Model / set | Toward correct option | Fixes wrong start |
|---|---|---|
| 27B · dev forms | 50.5% | 44.9% |
| 27B · logic puzzles | 41.3% | 36.9% |
| 9B · dev forms | 56.0% | 43.8% |
| 9B · logic puzzles | 57.3% | 49.2% |

Gradients may still help once additional training is allowed. In separate, unpublished work, a 4B model fine-tuned as an energy function [11] reaches 94.5% accuracy on 289 held-out puzzles with gradient descent plus search, compared with 75.1% without gradients. Since that experiment uses additional training, it falls outside the frozen-model results reported here.

## C.2 Rewriting the question

We also let DeepSeek V4.1 Flash rewrite the verification question using GEPA [1], keeping the rewrites that scored better on training tasks. On Qwen3.5-9B, this raised validation accuracy from 37.8% to 42.2%. On held-out sets, however, the rewrite helped only slightly on the logic puzzles and one internal set and hurt on dev and held-out forms, so everything in the main text uses the plain question.